\pdfoutput=1

\documentclass[11pt]{article}

\usepackage[]{ACL2023}

\usepackage{times}
\usepackage{latexsym}
\usepackage{microtype}

\usepackage{xcolor}
\usepackage{graphicx}

\usepackage{booktabs}
\usepackage{multirow}
\usepackage{hhline}
\usepackage{subcaption}

\usepackage{tabularx}
\usepackage{arydshln}
\usepackage{pifont}
\usepackage{amsmath}
\usepackage{hyperref}

\usepackage[T1]{fontenc}

\usepackage[utf8]{inputenc}

\usepackage{microtype}

\usepackage{inconsolata}

\title{SEE: Continual Fine-tuning with Sequential Ensemble of Experts}

\author{ 
 Zhilin Wang$^{\clubsuit }$\hspace{0.5mm},
 Yafu Li$^{\spadesuit}$\hspace{0.5mm},
 Xiaoye Qu$^{\spadesuit}$\hspace{0.5mm}, 
 Yu Cheng$^{\heartsuit}$\hspace{0.5mm}\hspace{0.5mm}\\
 \quad$^\clubsuit$Jilin University \ \ \ \quad$^\spadesuit$ Shanghai AI Laboratory\\ \quad$^\heartsuit$Chinese University of Hong Kong \\
 \texttt{\{linzwcs,yafuly\}@gmail.com}, \texttt{quxiaoye}@pjlab.org.cn  \\
 \quad\texttt{chengyu@cse.cuhk.edu.hk} \\
}

\begin{document}
\maketitle
\begin{abstract}


Continual fine-tuning of large language models (LLMs) suffers from catastrophic forgetting. 
Rehearsal-based methods mitigate this problem by retaining a small set of old data. 
Nevertheless, they still suffer inevitable performance loss.
Although training separate experts for each task can help prevent forgetting, effectively assembling them remains a challenge. 
Some approaches use routers to assign tasks to experts, but in continual learning, they often require retraining for optimal performance.
To address these challenges, we introduce the Sequential Ensemble of Experts (SEE) framework. 
SEE removes the need for an additional router, allowing each expert to independently decide whether a query should be handled. 
The framework employs distributed routing, and during continual fine-tuning, SEE only requires the training of new experts for incoming tasks rather than retraining the entire system.
Experiments reveal that the SEE outperforms prior approaches, including multi-task learning, in continual fine-tuning. 
It also demonstrates remarkable generalization ability, as the expert can effectively identify out-of-distribution queries, which can then be directed to a more generalized model for resolution.
This work highlights the promising potential of integrating routing and response mechanisms within each expert, paving the way for the future of distributed model ensembling. The code is available 
\href{https://github.com/Linzwcs/SEE}{here}.




\end{abstract}

\section{Introduction}

Large language models (LLMs) have demonstrated impressive performance across various scenarios~\cite{touvron2023llama,openai2024gpt4technicalreport}. However, they often require further refinement or the ability to learn new tasks for real-world applications. In these scenarios, LLMs are typically trained through a sequence of tasks performed successively, a process known as continual fine-tuning~\cite{luo_empirical_2024}. As a subclass of continual learning, continual fine-tuning also suffers catastrophic forgetting, whereby LLMs tend to lose previously acquired knowledge as they assimilate new information~\cite{kirkpatrick_overcoming_2017,wu_continual_2024,shi_continual_2024,qu2025survey}.

\begin{figure}[t]
    \centering
    \includegraphics[width=\linewidth]{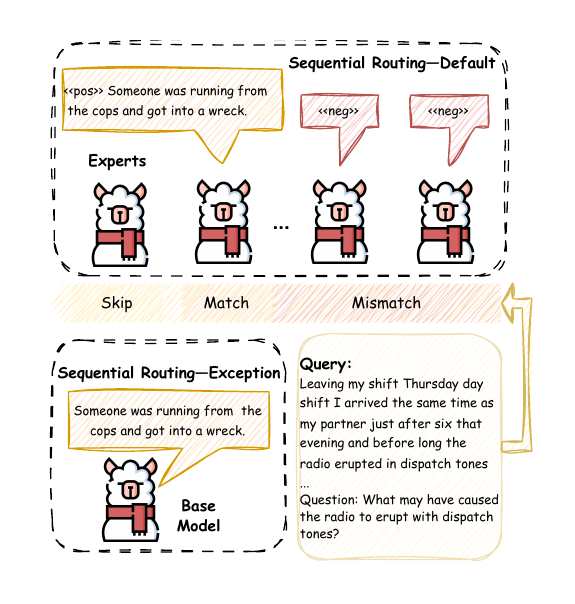}
    \vspace{-10mm}
    \caption{The inference process in the SEE framework involves a query being passed sequentially through a series of experts until it matches one, which then generates a response. If an expert fails to produce a special indicator, the query is routed to the base model, which is considered to possess the best generalization ability.}
    \label{fig:intro}
\end{figure}

While there have been many studies on alleviating catastrophic forgetting, rehearsal-based methods prove to be the most effective way in continual fine-tuning~\cite{zhang_citb_2023}. 
These methods work by combining new data with a subset of previous data~\cite{de_masson_d_autume_episodic_2019,rolnick_experience_2019}, or synthesized instances from the prior data distribution~\cite{huang_mitigating_2024}.
However, despite their effectiveness in preserving prior knowledge, these methods still experience some inevitable performance loss.

Although training separate experts for each task can prevent forgetting, organizing these experts effectively remains a challenge. 
The methods of organizing multiple models to create a more powerful model are commonly called Mixture of Experts (MoE) \cite{zhu2024llama,qu2024llama}.
However, these methods typically operate on a fixed set of tasks and domains~\cite{lu_routing_2023, jiang_llm-blender_2023} and introduce additional routers to manage the models~\cite{jang_exploring_2023,lu2024twin}. 
Except for their inefficiency in incremental scenarios, recent work~\cite{lv2025autonomyofexpertsmodels} also suggests that the separations between the router's decision-making and experts' execution may lead to suboptimal expert selection and ineffective learning.


To this end, we propose a framework referred to as the \textbf{S}equential \textbf{E}nsemble of \textbf{E}xperts (SEE), which combines the rehearsal-based method with MoE.
SEE comprises a base model and a sequence of experts, each proficient in recognizing questions within its specialized domain and generating precise responses accordingly.
These experts are derived from the base model, such as Llama2, and are further specialized for specific tasks through LoRA training.
When a new task arrives, SEE combines the new data with a small subset of instances from previous tasks, similar to rehearsal-based methods, and reconstructs the dataset to generate both positive and negative instances.
A new expert is then trained through supervised fine-tuning (SFT) on these instances and appended to the end of the expert sequence.
SEE integrates all experts via a distributed routing mechanism called sequential routing, as illustrated in Figure~\ref{fig:intro}.

Empirical experiments on task sequences from the SuperNI dataset show that SEE outperforms previous rehearsal-based methods and matches or exceeds multi-task learning (MTL) performance.
Besides, SEE further outperforms rehearsal-based methods on MMLU~\cite{hendryckstest2021} and exhibits astonishing knowledge preservation and out-of-distribution (OOD) generalization ability.
Moreover, when regarded as a single model, the perplexity of SEE proves lower than MTL on tasks during continual learning. 
Additionally, while SEE uses special indicators to identify a query, our analysis demonstrates that introducing additional tokens as indicators is more efficient than using tokens from the vocabulary for constructing indicators.
Although SEE can theoretically utilize samples not included in the continual learning process to construct negative instances, this would significantly degrade routing accuracy, thus highlighting the essence of rehearsal. Finally, we demonstrate that the additional latency introduced by sequential routing is affordable.



\section{Preliminaries}

\paragraph{Continual Fine-tuning}  

Let \( \mathcal{M} \) denote the LLM. The model undergoes continual fine-tuning across \( N \) stages, with each stage \( i \) involving updates based on the instruction dataset $D^{(i)}$ corresponding to task \( T_{i} \)~\cite{luo_empirical_2024}. Formally, the update process at each stage is defined as:
\begin{equation}
\mathcal{M}_{i} = \text{Update}(\mathcal{M}_{i-1}, D^{(i)})    
\end{equation}
where \( \mathcal{M}_0 \) denotes the initial model.

\paragraph{Rehearsal-based Methods}   As extensively discussed in the literature \cite{de_masson_d_autume_episodic_2019, rolnick_experience_2019}, 
Rehearsal-based methods involve sampling a subset of instances from earlier stages to expand the training data for the current stage. Formally, the augmented training set is expressed as:  
\begin{equation}
D^{(t)} \cup \sum_{i=1}^{t-1} \left( \tau D^{(i)} \right)    
\end{equation}
where \( \tau \) denotes the rehearsal ratio, representing the proportion of training instances sampled from previous stages. At stage \( i \), the model \( \mathcal{M}_{i} \) is updated as follows: \begin{equation}
\mathcal{M}_{i} = \text{Update}(\mathcal{M}_{i-1}, D^{(t)} \cup \sum_{i=1}^{t-1} \left( \tau D^{(i)} \right))    
\end{equation}
The rehearsal-based methods can effectively mitigate the issue of catastrophic forgetting in LLMs, as demonstrated in prior studies \cite{scialom-etal-2022-fine, mok-etal-2023-large}.

\section{Methods}

\begin{figure*}[!ht]
    \centering
    \includegraphics[width=\linewidth]{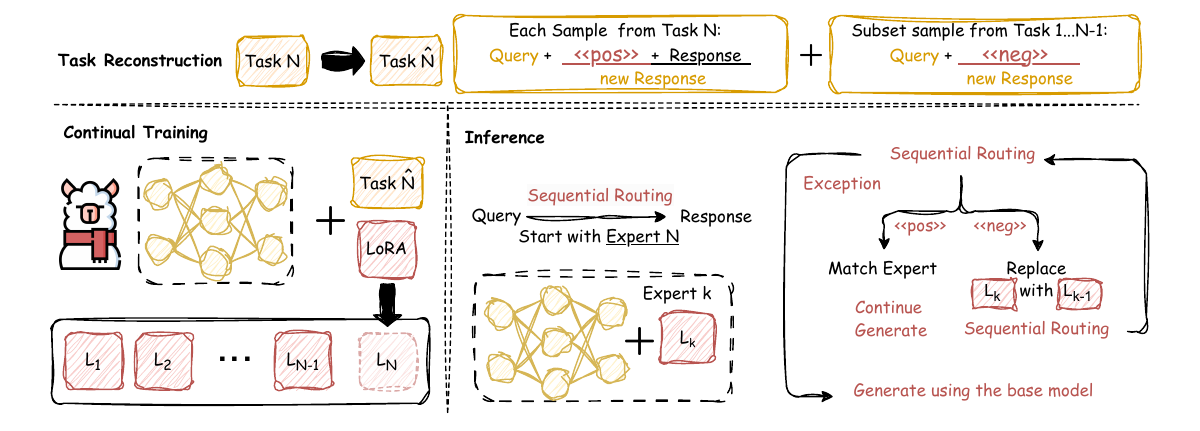}
    \caption{
   \textbf{ Overview of the SEE Framework:}
    The SEE framework operates in three steps when a new task is introduced:  
    (1) Task Reconstruction:  Current data are combined with sampled instances from previous tasks to guide expert routing and responses.
    (2) SFT: A new expert is trained using a new LoRA on the reconstructed task.  
    (3) Inference: All experts are integrated into a MoE system through sequential routing, enabling powerful inferences by leveraging the entire system.
}
\label{fig:SEE-framework}
\vspace{-2mm}
\end{figure*}

In contrast to previous research, our work integrates the concept of MoE with rehearsal-based methods. For each task $T_i$, we assign a specific expert $E_i$ to handle it. These experts are able to assess whether a query falls within their scope of responsibility. If so, they proceed to generate the response. SEE integrates the experts and the base model through a process of sequential routing. 
An overview of SEE is shown in Figure~\ref{fig:SEE-framework}, and it can be divided into three main steps: (1) Task Reconstruction, (2) Expert Training, and (3) Inference.


\paragraph{Task Reconstruction}
At stage \(i\), we reconstruct each task dataset \(D_i\) to enable the experts to clarify their responsibilities. 
Initially, each sample \(x\) in \(D_i\) can be represented as a tuple \((q, r)\), where \(q\) is the query and \(r\) is the corresponding response. 
We transform each \(x\) into \(\hat{x}\), which consists of the tuple \((q, o_{pos}, r)\), by adding a special positive indicator \(o_{pos}\). 
Additionally, we sample \(\tau\)\% of the data from previous tasks, as in rehearsal-based methods, but ignore the response \(r\) and add a negative indicator \(o_{neg}\). The final constructed \(\hat{D}_i\) consists of positive samples \(\{(q, o_{pos}, r)_j\}_{j=1}^n\) from the current task and a small portion of negative samples \(\{(q, o_{neg})_j\}_{j=1}^m\) from previous tasks:
\begin{equation}
   \hat{D}_i=\{(q, o_{pos}, r)_j\}_{j=1}^n \cup  \{(q, o_{neg})_j\}_{j=1}^m    
\end{equation}

\paragraph{SFT} 
We frame both the decision-making and generation processes of the experts within the paradigm of standard supervised fine-tuning. Multiple LoRA adapters are used as expert weights, enabling the base model \(\mathcal{M}_{0}\) to specialize into task-specific experts.
Starting with the initial model \(\mathcal{M}_0\), a new LoRA adapter \(\mathcal{L}_i\) is introduced whenever a new task \(T_i\) arises. The expert for this task is represented as \(E_{(\mathcal{M}_0, \mathcal{L}_i)}\). 
Next, we obtain the dataset \(\hat{D}_i\) and maximize the following optimization goal:
\begin{equation}
    \begin{aligned}
        \max_{\mathcal{L}_i} \frac{1}{n+m} \ &\left( \sum_{i=1}^{n} E_{(\mathcal{M}_{0}, \mathcal{L}_i)} \left( r^{(i)}, o_{pos}^{(i)} \mid q^{(i)} \right) \right. \\
        + & \left. \sum_{i=1}^{m} E_{(\mathcal{M}_{0}, \mathcal{L}_i)} \left( o_{neg}^{(i)} \mid q^{(i)} \right) \right)
    \end{aligned}
\end{equation}

Given a query, the expert first determines whether the question falls under its responsibility and decides whether to provide an answer. By sequentially ensembling all experts, we can handle various tasks through sequential routing. 

\paragraph{Inference }
After continuing fine-tuning on \( N \) tasks, we obtain \( N \) experts, denoted as \( (E_{(\mathcal{M}_{\theta}, \mathcal{L}_1)}, \dots, E_{(\mathcal{M}_{\theta}, \mathcal{L}_N)}) \), where the \( i \)-th expert specializes in task \( T_i \).
When a question is posed, the expert  $E_{(\mathcal{M}_{\theta}, \mathcal{L}_N)}$ generates an indicator to decide whether to continue generating a response. If the indicator is $o_{neg}$, the expert  $E_{(\mathcal{M}_{\theta}, \mathcal{L}_N)}$ stop generating and routes the query to the expert $E_{(\mathcal{M}_{\theta}, \mathcal{L}_{N-1})}$
We iteratively repeat this process until an expert generates \( o_{pos} \). Once this happens, the expert is chosen to generate the final response.
If an expert fails to produce an indicator, we route the query to the base model $\mathcal{M}_0$ and let it to answer the question. 
This entire routing process is referred to as Sequential Routing.
By applying sequential routing, we combine all individual experts into a more powerful model.

\section{Experiments}

\subsection{Setup}
The setup of our experiments mainly follows the paper~\cite{huang_mitigating_2024}.

\paragraph{Dataset} Our experiments are conducted using the SuperNI~\cite{wang-etal-2022-super} dataset, a comprehensive and extensive benchmark designed for instruction tuning. To simulate a typical continual learning process, we select a subset of 10 tasks from SuperNI, representing a range of domains and tasks. Each task includes 2,000 instances for training and 500 instances for evaluation. Additional details can be found in Appendix~\ref{app:task-info} or the referenced paper~\cite{huang_mitigating_2024}. For simplicity, the default continual learning order for the \{5, 10\} SuperNI tasks is as follows: QA $\rightarrow$ QG $\rightarrow$ SA $\rightarrow$ Sum. $\rightarrow$ Trans. ($\rightarrow$ DSG $\rightarrow$ Expl. $\rightarrow$ Para. $\rightarrow$ PE $\rightarrow$ POS). 

\paragraph{Base LLMs}
Following \cite{huang_mitigating_2024}, we utilize three base LLMs: Llama-2-7B~\cite{touvron2023llama}, Llama-2-7B-Chat~\cite{touvron2023llama}, and Alpaca-7B~\cite{alpaca} in our experiments. However, we also provide the results of the latest models, including Llama3.1-8B, Qwen2.5-7B, and Mistral-7B-v0.3, in Appendix~\ref{app:latest_model}.

\paragraph{Baselines} We primarily compare the SEE method with the following baselines:
\begin{itemize}
    \item \textbf{Multi-task Learning (MTL)}: The most commonly used baseline, where all tasks are trained simultaneously. 
    \item \textbf{Average Single-task Learning (AvgSTL)}: For each task, AvgSTL trains a dedicated model and assesses its performance. The final results are derived by averaging the performance across all tasks. 
    \item \textbf{Non-rehearsal}: A naive baseline where the LLM is fine-tuned with only the instruction data $T^{(t)}$ at each stage $t$.
    \item \textbf{RandSel($\tau$)}~\cite{scialom-etal-2022-fine}: A baseline where we randomly sample $\tau = \{1, 10\}\%$ of the original instruction data for each previous task. As noted in~\cite{scialom-etal-2022-fine}, the capabilities of language models can be effectively preserved with $\tau = 1\%$.
\end{itemize}
Additionally, we include the \textbf{KmeansSel}~\cite{huang_mitigating_2024} and \textbf{SSR}~\cite{huang_mitigating_2024} methods in our experiments.

\paragraph{Evaluation Metrics}
Following~\citet{huang_mitigating_2024}, we evaluate LLM performance on SuperNI tasks using the ROUGE-L metric~\cite{lin-2004-rouge}, which aligns closely with human evaluation~\cite{wang-etal-2022-super}. The following ROUGE-L-based metrics are selected for our experiments, where $a^{(i)}_t{j}$ denotes the ROUGE-L score for task $j$ at stage $i$:

\begin{itemize}
    \item \textbf{Average ROUGE-L (AR)}: The average LLM performance across all $T$ tasks at stage $T$:
    \begin{equation}
        \mathbf{AR} = \frac{1}{T}\sum_{i=1}^{T}a^{(T)}_{i}
    \end{equation}
    \item \textbf{Backward Transfer (BWT)}: Assesses the impact of learning new tasks on previous ones by comparing final performance $a^{(T)}_{i}$ to performance during stage $i$:
    \begin{equation}
        \mathbf{BWT} = \frac{1}{T-1}\sum_{i=1}^{T-1}(a^{(T)}_{i} - a^{(i)}_{i})
    \end{equation}
    Negative BWT indicates catastrophic forgetting of previously learned knowledge.
\end{itemize}

On the other hand, we do not use Forward transfer metric~\cite{lopez2017gradient} to evaluate generalization abilities. In our design, this ability is tied to the base model. Thus, We employ a different instruction tuning dataset and evaluate how effectively it can be routed to the base model.

\paragraph{Training Details}
We set the LoRA rank to 8 and the dropout rate to 0.1. The Adam optimizer is used with an initial learning rate of \(2 \times 10^{-4}\). The global batch size is 32 in our all experiments. The input and output lengths are configured to 1,024 and 512, respectively. Each LLM is trained for 3 epochs, with evaluation conducted using the final checkpoint.

\subsection{Results on 5 SuperNI Tasks}
\begin{table*}[!ht]
\centering
\small
\tabcolsep=11pt
\begin{tabular}{lcccccccccccc}
\toprule
\multirow{2}{*}{\textbf{Model}}
& \multicolumn{2}{c}{\textbf{Order 1}} & \multicolumn{2}{c}{\textbf{Order 2}} & \multicolumn{2}{c}{\textbf{Order 3}} & \multicolumn{2}{c}{\textbf{Avg.}} \\
\cmidrule(lr){2-3} \cmidrule(lr){4-5} \cmidrule(lr){6-7} \cmidrule(lr){8-9}
&  \textbf{AR} & \textbf{BWT} &  \textbf{AR} & \textbf{BWT} &  \textbf{AR} & \textbf{BWT} & \textbf{AR} & \textbf{BWT} \\
\midrule
\multicolumn{9}{c}{\textit{Llama-2-7B}} \\
\midrule
AvgSTL & 52.95 & - & 52.95 & - & 52.95 & - & 52.95 & -\\
MTL & 53.05 & - & 53.05 & - & 53.05 & - & 53.05  & -\\
Non-rehearsal & 17.67 & -44.09 & 15.25 & -47.09 & 24.16 & -35.99 & 19.03  & -42.39 \\
\hdashline
RandSel(1\%) & 51.16 & -2.34 & 49.21 & -4.36 & 48.63 & -5.37 & 49.67  & -4.02 \\
KMeansSel(1\%) & 50.20 & -3.12 & 49.75 & -4.11 & 50.12 & -3.61 & 50.02  & -3.61 \\
RandSel(10\%) & 50.81 & -2.32 & 50.04 & -3.31 & 50.11 & -3.42 & 50.32  & -3.02 \\
KMeansSel(10\%) & 50.44 & -3.03 & 50.61 & -2.32 & 49.89 & -3.53 & 50.31  & -2.96 \\
SSR & 52.61 & -0.23 & 51.70 & -1.22 & 52.16 & -0.93 & 52.16\  & -0.79 \\
\hdashline
SEE(1\%) & 52.86 & -0.47 & \textbf{53.19} & -0.38 & 53.37 & -0.13 & 53.14  & -0.33\\
SEE(10\%) & \textbf{53.40} & \textbf{-0.00} & 52.94 & \textbf{-0.01} & \textbf{53.44} & \textbf{-0.00} & \textbf{53.26} & \textbf{-0.00} \\
\midrule
\multicolumn{9}{c}{\textit{Llama-2-7B-chat}} \\
\midrule
AvgSTL & 52.13 & - & 52.13 & - & 52.13 & - & 52.13 & -\\
MTL & 52.81 & - & 52.81 & - & 52.81 & - & 52.81  & -\\
Non-rehearsal & 23.87 & -36.31 & 30.96 & -27.41 & 42.06 & -13.50 & 32.30  & -25.74 \\
\hdashline
RandSel(1\%) & 51.28 & -1.96 & 49.77 & -3.70 & 49.41 & -4.29 & 50.15  & -3.32 \\
KMeansSel(1\%) & 51.82 & -1.25 & 50.71 & -2.44 & 50.22 & -3.42 & 50.92  & -2.37 \\
RandSel(10\%) & 50.59 & -2.57 & 50.72 & -2.45 & 50.24 & -2.87 & 50.52  & -2.63 \\
KMeansSel(10\%) & 50.81 & -2.55 & 51.39 & -1.42 & 50.22 & -2.84 & 50.81  & -2.27 \\
SSR & 52.52 & -0.23 & 52.49 & -0.35 & 52.73 & 0.05 & 52.58  & -0.18 \\
\hdashline
SEE(1\%) & 52.85 & -0.09 & 52.77 & -0.04 & 53.03 & -0.05 & 52.88 & -0.06 \\
SEE(10\%) & \textbf{53.13} & \textbf{-0.00} & \textbf{53.15} & \textbf{-0.00} & \textbf{52.93} & \textbf{-0.01} & \textbf{53.07}  & \textbf{-0.00}\\
\midrule
\multicolumn{9}{c}{\textit{Alpaca-7B}} \\
\midrule
AvgSTL & 51.78 & - & 51.78 & - & 51.78  & - & 51.78 & -\\
MTL & 52.79 & - & 52.79 & - & 52.79 & - & 52.79  & -\\
Non-rehearsal & 17.24 & -44.21 & 45.40 & -9.03 & 35.60 & -21.45 & 32.75  & -24.90 \\
\hdashline
RandSel(1\%) & 51.61 & -0.93 & 49.08 & -4.68 & 49.01 & -4.85 & 49.90  & -3.49 \\
KMeansSel(1\%) & 51.37 & -1.53 & 50.53 & -2.68 & 50.15 & -3.17 & 50.68  & -2.46 \\
RandSel(10\%) & 50.91 & -1.82 & 50.88 & -2.11 & 49.98 & -3.59 & 50.59  & -2.51 \\
KMeansSel(10\%) & 50.78 & -2.05 & 51.20 & -1.76 & 49.76 & -3.48 & 50.58  & -2.43 \\
SSR & 
\textbf{52.52} & -0.14 & 51.74 & -1.21 & 52.33 & -0.51 & 52.20 & -0.62 \\
\hdashline
SEE(1\%) & 52.50 & \textbf{-0.00} & \textbf{52.71} & -0.01 & 52.44 & \textbf{-0.00} & 52.55  &\textbf{-0.00} \\
SEE(10\%) & 52.50 & \textbf{-0.00} & 52.66 & \textbf{-0.00} & \textbf{52.55} & \textbf{-0.00} & \textbf{52.57} & \textbf{-0.00} \\
\bottomrule
\end{tabular}
\caption{Final results on 5 SuperNI tasks under 3 types of continual learning orders. More details about task orders can be found in Appendix \ref{app:five-task}. The results of the latest models, including Llama3.1-8B, Qwen2.5-7B, and Mistral-7B-v0.3, are presented in Appendix~\ref{app:latest_model}.}
\label{tab:superni_5}
\end{table*}


We present the experimental results on five SuperNI tasks in Table~\ref{tab:superni_5}. 
The table clearly demonstrates that SEE methods consistently outperform previous rehearsal-based approaches in both AR and BWT metrics across nearly all model configurations and task sequences, even when only 1\% of queries from previous tasks are retained. 
Moreover, the additional parameters enable SEE to not only compete with, but potentially surpass MTL in the AR metric—traditionally considered the upper bound for continual learning in a single LLM. 
In contrast, AvgSTL exhibits lower performance compared to MTL. This trend aligns with previous research~\cite{caruana1997multitask,ruder2017overviewmultitasklearningdeep}, which suggests that models trained on multiple tasks simultaneously can improve generalization and outperform those trained on tasks in isolation.

While higher BWT scores generally indicate reduced catastrophic forgetting, for the SEE model, which does not inherently 'forget' knowledge, they reflect improved accuracy in routing instances. As shown in Table~\ref{tab:superni_5}, even with a limited number of instances from previous tasks, the SEE model demonstrates its ability to effectively learn to route queries. This success can be attributed to the distinct patterns often present in the instructions of different tasks. Consequently, the SEE model proves to be particularly well-suited for applications in Continued Fine-tuning.

The value of \(\tau\) plays a crucial role in both the performance of each expert within the SEE model and the extent of forgetting across the entire system. As illustrated in Table~\ref{tab:superni_5}, the BWT values for the SEE(1\%) model are lower than those for SEE(10\%), suggesting that providing a greater number of instances from previous tasks helps SEE develop a more effective routing strategy. However, while SEE(10\%) achieves near 'zero forgetting,' its AR values do not show as significant an improvement as those of SEE(1\%), and in certain cases, they even experience a slight decline. This indicates that incorporating more instances from previous tasks may negatively impact the performance of experts on their respective tasks.

\subsection{Results on 10 SuperNI Tasks}

We further investigate the performance of SEE in continuous learning across 10 tasks sequentially, and the results based on Llama2 are presented in Table~\ref{tab:superni_10}.
Despite the increase in tasks, the table shows that SEE outperforms previous rehearsal-based methods in both AR and BWT metrics.
Besides, in the AR metric, SEE(1\%) proves to be competitive with MTL, while SEE(10\%) delivers a 1.5\% improvement over the MTL baseline.


\begin{table}[t]
    \centering
    \small
    \tabcolsep=8pt
    \begin{tabular}{lccc}
    \toprule
    \textbf{Model}& \textbf{AR}  & \textbf{BWT} \\
    \midrule
    \multicolumn{4}{c}{\textit{Llama-2-7b}} \\\midrule
     AvgSTL
    &  65.63  & - \\
    MTL
    & 64.69  & - \\
    Non-rehearsal 
    & 17.33 & -53.64 \\
    \hdashline
    RandSel(1\%)
    & 60.64  & -5.69 \\
    KMeansSel(1\%) & 60.51& -5.39 \\
    RandSel(10\%)
    & 61.49  & -4.03 \\
    KMeansSel(10\%) & 60.93  & -3.90 \\
    SSR & 63.23  & -1.56 \\
    \hdashline
    SEE (1\%) &  64.64  & -0.56  \\
    SEE (10\%) & \textbf{66.19} & \textbf{--0.00}  \\
    \bottomrule
    \end{tabular}
    \caption{Final results for Llama-2-7B on 10 SuperNI tasks. The results of the latest models are presented in Appendix~\ref{app:latest_model}.}
    \label{tab:superni_10}
\end{table}
\begin{figure}
    \centering
    \includegraphics[width=\linewidth]{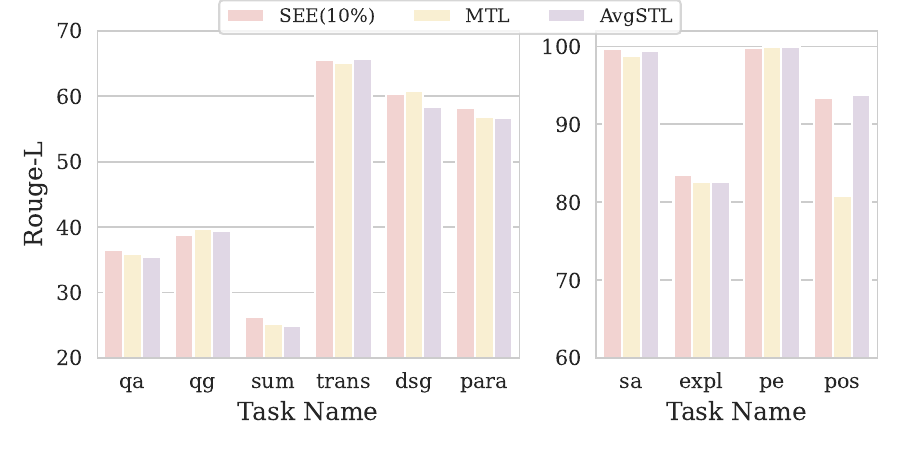}
    \vspace{-8mm}
    \caption{The ROUGE-L scores for SEE (10\%), MTL, and AvgSTL across the 10 SuperNI tasks are presented in two figures, separated according to the magnitude of the values for clearer comparison.}
    \label{fig:superni_10_specific}
\end{figure}
Comparing the BWT values in Table~\ref{tab:superni_5} and Table~\ref{tab:superni_10}, it is evident that increasing the number of tasks leads to a decrease in the BWT value, indicating a greater challenge in retaining knowledge. Despite this, SEE(10\%) still achieves “zero forgetting”.
It demonstrates the potential of SEE to effectively extend to continuous learning with a larger number of tasks.

Figure~\ref{fig:superni_10_specific} presents an analysis of the performance across individual tasks. This figure compares the ROUGLE-L scores of SEE(10\%), MTL, and AvgSTL. As shown, MTL demonstrates lower performance on the sa and pos tasks, both of which are classification tasks. This may be due to task conflicts~\cite{javaloy2022rotograd,mueller-etal-2022-text}, where learning one task negatively impacts performance on another.
Additionally, SEE(10\%) outperforms AvgSTL in most of the generation tasks, highlighting that the introduction of routing capabilities enhances the generative performance for these tasks.

\section{Analysis}
\label{sec:analysis}

\begin{figure}[h]
    \centering
    \includegraphics[width=\linewidth]{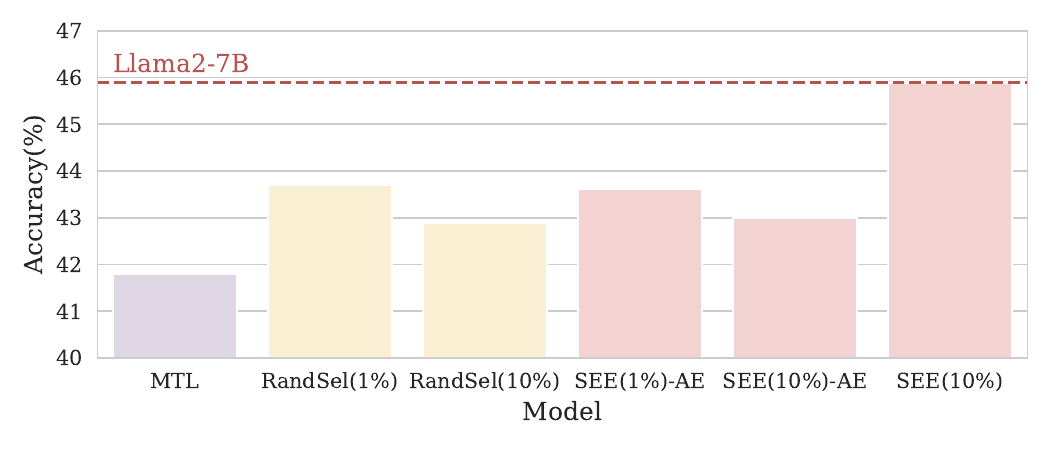}
    \vspace{-8mm}
    \caption{The performance of different methods after continuous learning of 10 SuperNI tasks on the MMLU benchmark. SEE(10\%)-AE and SEE(1\%)-AE represent the average performance of experts in SEE.}
 
    \label{fig:icl-analysis}
\end{figure}

\paragraph{Generalization}
\label{subsec:Generalization}
While the SEE model has demonstrated outstanding performance in tasks during continuous fine-tuning, it is also essential to assess its generalization.
In our design, the generalization ability of the SEE depends on two key factors: the average generalization ability of the experts and the accuracy with which the SEE routes out-of-distribution (OOD) queries to the base model. 
The former refers to the scenario where an OOD query is incorrectly routed to an expert, in which case SEE's generalization ability is determined by the average generalization ability of the experts. The latter reflects SEE's ability to correctly identify OOD tasks.
To assess this, we evaluate SEE’s average expert performance on MMLU~\cite{hendryckstest2021} and the standard performance of SEE. We compare their performance against that of the base model, MTL, and RandSel. The experiment is based on Llama-2-7B and the results are presented in Figure~\ref{fig:icl-analysis}.



As shown in Figure~\ref{fig:icl-analysis}, the base model $\mathcal{M}_{\theta}$ achieves the highest accuracy. The average performance of the experts in SEE is similar to that of RandSel. 
However, It is important to consider the OOD routing accuracy when evaluating the actual performance of SEE. Specifically, 99.72\% of instances are routed to the base model in SEE(10\%), while this percentage is 99.52\% in SEE(1\%). As a result, the actual performance of the SEE framework significantly exceeds that of other methods, demonstrating its superior generalization ability.

\begin{figure}[b]
    \centering
    \includegraphics[width=\linewidth]{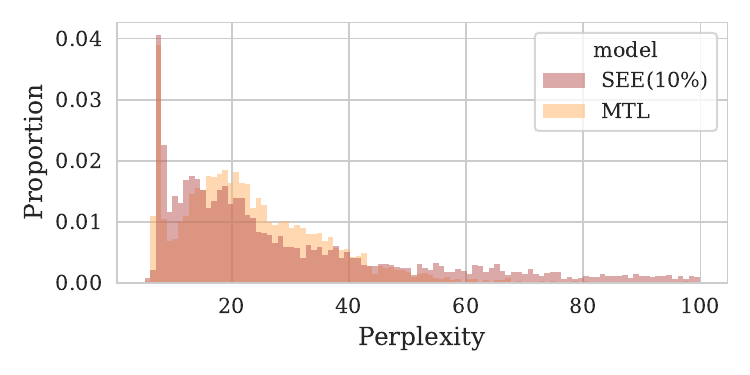}
    \vspace{-8mm}
    \caption{Comparison of the perplexity distribution between SEE and MTL across 10 SuperNI tasks.}
    \label{fig:perplexity}
\end{figure}

\paragraph{Perplexity across 10 SuperNI tasks}
Treating SEE as a unified model, we compute the perplexity for each instance \((q, r)\). First, the instance is transformed into \((q, o_{pos}, r)\), where \(o_{pos}\) represents a positive indicator. This transformed instance is then routed to the relevant expert within SEE. Then, the perplexity is computed by the selected expert.
We evaluate the perplexity of SEE(10\%) across 10 SuperNI tasks and compare the results with those of MTL, which, in other hand,  computes the perplexity using the original \((q,r)\) pairs.
Figure~\ref{fig:perplexity} illustrates that SEE achieves lower perplexity across the 10 SuperNI tasks, highlighting the effectiveness of the SEE framework.


\paragraph{Impact of \(\tau\) on Routing and Performance}

\begin{figure}[ht]
    \centering
    \includegraphics[width=\linewidth]{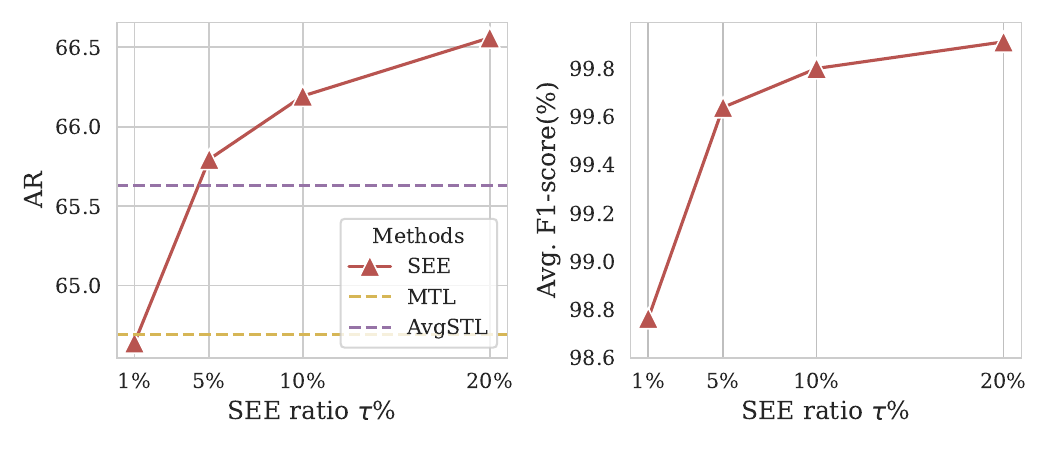}
    \vspace{-8mm}
    \caption{
        Impact of \(\tau\) on SEE: The left plot shows how the AR metric increases as \(\tau\) grows, while the right illustrates the improvement in routing accuracy.
    }
    \label{fig:tau-analysis}
\end{figure}

To further investigate the impact of \(\tau\) on SEE, we introduce two additional SEE models with \(\tau\) values of 5\% and 20\% for continuous learning on the 10 SuperNI tasks. We then plot the variations in the AR metric and F1 score of routing with \(\tau\), and the results are shown in Figure~\ref{fig:tau-analysis}. 

The figure shows that as \(\tau\) increases, both the AR metric and routing accuracy improve steadily. In particular, SEE(20\%) significantly outperforms both the MTL and AvgSTL methods. Moreover, SEE(20\%) achieves near-perfect routing. A more detailed analysis of the routing performance for each task can be found in Appendix~\ref{app:routing_analysis}. These results highlight the efficiency of SEE and its potential to be extended to a larger number of tasks.

\paragraph{Essence of Rehearsal}
The SEE framework utilizes sequential routing to ensemble all experts. Each expert must determine whether a question falls under its responsibility and decide whether to produce a response. While SEE samples negative instances from previous tasks to acquire this capability, it also brings up the question of whether using data from unrelated tasks (called pseudo-negative sampling) could have the same effect.
To test this, following the setup of SEE(10\%), we replace the negative samples from previous tasks with an equal number of instances from Alpaca-52k, perform the experiments on the 10 SuperNI tasks setup, and compare their performance before and after the changes.
Since the performance of SEE on OOD data mainly depends on the accuracy of instance routing to the base model (RB-Acc.), we evaluate the two versions of SEE on MMLU using this metric.

\begin{table}[h] 
\centering
\small
\begin{tabular}{lccc}
\toprule
\multirow{2}{*}{\textbf{Model}}  & {\textbf{MMLU}} & \multicolumn{2}{c}{\textbf{SuperNI}} \\
\cmidrule(lr){2-2}  \cmidrule(lr){3-3} \cmidrule(lr){3-4} 
 &  RB-Acc. & AR  & R-F1(\%) \\
\midrule
SEE(10\%)  &   99.72  &  66.19 & 99.80 \\
\qquad - w/pseudo  & 99.57  & 46.40 & 47.22 \\
\bottomrule
\end{tabular}
\caption{Performance comparison of SEE and SEE with pseudo-Negative sampling.}
\label{tab:pesudo-negative}
\end{table}

Table~\ref{tab:pesudo-negative} compares SEE(10\%) and its pseudo-negative sampling variant on the MMLU and 10 SuperNI tasks, using the Llama2-7B model. It shows that, without instances from previous tasks, the F1-score of routing (R-F1) drops significantly, leading to a substantial degradation in the AR metric. However, the performance of MMLU remains unchanged. These findings highlight the importance of rehearsal for SEE when solving tasks in continuous fine-tuning sequences.

\paragraph{Special token indicators}
While the SEE framework introduces additional tokens as special indicators, we further investigate whether these can be replaced with in-vocabulary tokens. To explore this, we propose two variants of special indicators: 1) Semantic Indicators, and 2) Non-Semantic Indicators. Semantic indicators are tokens that carry specific semantic meaning, whereas non-semantic indicators lack such meaning. In this experiment, we use "Yes" and "No" as semantic indicators, while "<<pos>>" and "<<neg>>" serve as non-semantic indicators. We evaluate the performance of these variants on 10 SuperNI tasks.
To further explore the impact of indicator type, we categorize the 10 tasks into two groups: Generation and  Classification.
As before, we set \(\tau\) to 10\% and use Llama2 as the base model.
The results are presented in Table~\ref{tab:special-token}.

\begin{table}[h]
\centering
\small
\begin{tabular}{lccc}
\toprule
\multirow{2}{*}{\textbf{Indicator Type}}    & {\textbf{Generation}} & \textbf{Classification} \\
\cmidrule(lr){2-2}  \cmidrule(lr){3-3} \cmidrule(lr){4-4} 
 &  AR & Acc.(\%) \\
\midrule
Additional   & 58.62      & 98.25  \\
No-Semantic   & 58.61  & 97.25  \\
Semantic  &  58.41 & 95.05 \\
\bottomrule
\end{tabular}
\caption{The performance of SEE(10\%) based on Llama2-7B across three indicator types in the 10 SuperNI tasks. The tasks are categorized into Generation and Classification.}
\vspace{-5mm}
\label{tab:special-token}
\end{table}

The table shows that indicator type primarily affects the performance of classification tasks, with minimal impact on generation tasks. In classification, the SEE framework with a semantic indicator yields the lowest accuracy, while SEE with a no-semantic indicator also performs worse than when additional indicators are used.
Therefore, introducing new special tokens as indicators to the vocabulary proves more effective for the SEE framework.

\paragraph{The Advantages of Combining Generation and Routing} 
Compared to using encoder-only models as routers, the expert in SEE framework excels in handling OOD scenarios. When a query does not correspond to a specific learning task, the experts typically avoid generating a specialized indicator. In such cases, SEE directs the query to the base model, regarded as the most generalized model in the system. Furthermore, this base model can be substituted with more powerful models, further improving the overall generalization capability of the SEE framework.
This approach, however, is not feasible with encoder-only models such as BERT. 
Moreover, integrating both generation and routing within a single expert not only reduces the parameter count but also streamlines the training pipeline, resulting in a more efficient and compact solution.

\paragraph{Extra Overhead of SEE}

While the sequential routing of SEE introduces additional overhead, we demonstrate that this cost remains entirely affordable. To quantify this, we calculate the additional latency introduced by the SEE framework compared to standard LLMs. It can be expressed as:
\begin{equation}
ExtraOverhead = \frac{1}{2} \times \frac{1}{1 + \frac{N}{M-1} \times \frac{TPOT}{TTFT}}    
\end{equation}

where \textbf{TTFT} denotes the \textit{Time to First Token}, representing the initial prefix processing latency; \textbf{TPOT} is the \textit{Time Per Output Token}, the average time required to generate each token; \textbf{N} is the total number of output tokens; and \textbf{M} is the number of experts.

As shown in previous work~\cite{zhong2024distservedisaggregatingprefilldecoding}, when serving an LLM with 13B parameters under a synthetic workload with an input length of 512 and an output length of $N = 64$ on an NVIDIA A100 (80GB), the ratio $\text{TPOT}/\text{TTFT}$ is approximately $1/10$. Assuming the presence of $M = 10$ experts, the additional overhead introduced by SEE is approximately $0.29$.

Even with a long input prefix and a short response, this results in only $0.29$ additional latency.
As the output length increases, this latency overhead can be reduced to a very low level. For instance, when the output length equals the input length (e.g., 512 tokens), the ratio $\text{TPOT}/\text{TTFT}$ exceeds $1/10$, and the additional overhead drops to less than $0.075$.
In many real-world scenarios, LLMs typically generate outputs significantly longer than the input, the relative overhead introduced by SEE can be expected to further decrease.
The detailed calculations can be found in Appendix~\ref{app:extra_overhead}.


\section{Related Work}
\paragraph{Continue learning}

Existing approaches to continue learning can be divided into regularization-based, architecture-based, and rehearsal-based methods.
Regularization-based methods constrain the inner distribution of LLMs to remain close to the original state by adding auxiliary loss or controlling parameter updates~\cite{kirkpatrick_overcoming_2017,cha_cpr_2021,huang_continual_2021,zhang_clle_2022}. However, these methods require manually tuned hyperparameters, limiting their applicability.
Rehearsal-based methods maintain a history buffer with a subset of previous datasets~\cite{de_masson_d_autume_episodic_2019, rolnick_experience_2019} or synthetic instances~\cite{huang_mitigating_2024}, replayed during future training. While effective at mitigating catastrophic forgetting, they cannot fully preserve prior knowledge.
Architecture-based methods introduce new parameters for each dataset and learn them independently, potentially avoiding knowledge loss~\cite{xu_reinforced_2018,huang_neural_2019,razdaibiedina_progressive_2023}. However, managing these additional parameters remains challenging.
In our framework, we use rehearsal with queries from previous tasks and organize additional parameters as expert weights, selecting them via sequential routing.


\paragraph{Mixture of experts}

The basic concept of MoE involves assembling an ensemble of experts to improve performance~\cite{hinton_moe_concept,jordan1994hierarchical}. Prior work on expert ensembles primarily focuses on fixed domains, tasks, and experts, achieving strong results~\cite{lu_routing_2023, jiang_llm-blender_2023}, but these methods are ill-suited for incremental scenarios or require additional routers~\cite{jang_exploring_2023}. Furthermore, \cite{lv2025autonomyofexpertsmodels} highlights the overlooked issue that separating the router's decision-making from the experts' execution may lead to suboptimal expert selection and ineffective learning. Recently, MoE has been applied to transformer architectures~\cite{dai_deepseekmoe_2024,jiang_mixtral_2024}, with models proposed to mitigate catastrophic forgetting. Building on GLaM~\cite{du_glam_2022}, Lifelong MoE~\cite{chen_lifelong_2023} expands experts incrementally while freezing previous ones. Some studies replace the standard MoE layer with LoRA-MoE for model editing~\cite{yang_moral_2024,wang_lemoe_2024} or to retain world knowledge in multi-task learning~\cite{dou_loramoe_2024}. However, these methods still struggle with catastrophic forgetting during continual fine-tuning. Our method, based on the MoE concept, sequentially assembles experts in continual fine-tuning, eliminating the need for an additional router and introducing a distributed routing mechanism called sequential routing.

\section{Conclusion}

In this work, we propose the Sequential Ensemble of Experts (SEE), which is specifically designed to adapt to continuous fine-tuning. SEE resolves the separation between the router's decision-making and the experts' execution by integrating routing and response mechanisms within each expert. It employs a distributed routing method called sequential routing, improving the system's scalability. Our experiments demonstrate SEE’s effectiveness in handling both continual learning tasks and out-of-distribution instances, paving the way for future advancements in distributed model ensembling.

\section*{Limitations}
While SEE demonstrates remarkable performance in continual fine-tuning, the increasing number of parameters as the task count grows calls for further consideration. Additionally, similar to rehearsal-based approaches, the volume of rehearsal data also expands proportionally with the number of tasks. These challenges are not unique to SEE and have been acknowledged by other methods, underscoring the necessity for continued advancements in this area.

\bibliography{anthology,custom}
\bibliographystyle{acl_natbib}


\appendix

\section{Details of the Selected 10 SuperNI Tasks}

\label{app:task-info}
Table \ref{tab:task-info} provides an overview of the 10 SuperNI tasks selected for our primary experiments. To simplify the discussion, we use abbreviations to refer to these tasks throughout the paper. The SuperNI dataset can be accessed at \url{https://github.com/allenai/natural-instructions} for further reference.

\section{More Details of Experiments on 5 SuperNI Tasks}

\label{app:five-task}
We assess the performance of various methods across 5 SuperNI tasks, each with different continuous fine-tuning sequences. Table \ref{tab:cl-order} outlines the 3 specific continual learning orders for the five SuperNI tasks used in our experiments.

\begin{table}[h]
    \tabcolsep=7pt
    \small
    \centering
    \begin{tabular}{cp{0.5cm}c}
        \toprule
        \textbf{Order} & & \textbf{Task Sequence} \\
        \midrule
         1 & & QA $\rightarrow$ QG $\rightarrow$ SA $\rightarrow$ Sum. $\rightarrow$ Trans. \\
         2 & & Trans. $\rightarrow$ SA $\rightarrow$ QA $\rightarrow$ Sum. $\rightarrow$ QG \\
         3 & & Sum. $\rightarrow$ QG $\rightarrow$ Trans. $\rightarrow$ QA $\rightarrow$ SA \\
         \bottomrule
    \end{tabular}
    \caption{Continual learning orders on 5 SuperNI tasks.}
    \label{tab:cl-order}
\end{table}

\section{Results of the Latest Models}
\label{app:latest_model}
To demonstrate the efficiency of our method on the latest models, we present the AR scores of the Qwen2.5-7B, Mistral-7B-v0.3, and Llama3.1-8B models. We evaluate them in both 5-task and 10-task settings, with the results shown in Table~\ref{tab:latest_5} and Table~\ref{tab:latest_10}.

\begin{table*}[h]
\centering
\begin{tabular}{lcccccc}
\hline
\multirow{2}{*}{\textbf{Methods}} & \multicolumn{2}{c}{\textbf{Llama3.1-8B}} & \multicolumn{2}{c}{\textbf{Mistral-7B-v0.3}} & \multicolumn{2}{c}{\textbf{Qwen2.5-7B}} \\
\cline{2-7}
                               & \textbf{Base} & \textbf{Instruct} & \textbf{Base} & \textbf{Instruct} & \textbf{Base} & \textbf{Instruct} \\
\hline
\textbf{AvgSTL}            & 51.64 & 52.19 & \textbf{52.38} & 51.96 & 51.37 & 51.43 \\
\textbf{MTL}                 & 51.67 & 51.92 & 51.77 & \textbf{52.33} & \textbf{51.38} & 51.38 \\
\textbf{SEE(1\%)}             & 51.97 & 52.05 & 51.48 & 51.76 & 51.19 & 51.65 \\
\textbf{SEE(10\%)}            & \textbf{52.14} & \textbf{52.30} & 51.74 & 52.31 & 51.36 & \textbf{51.80} \\
\hline
\end{tabular}
\caption{AR results of the latest models on five SuperNI tasks in the default order.}
\label{tab:latest_5}
\end{table*}

\begin{table*}[h]
\centering
\begin{tabular}{lcccccc}
\hline
\multirow{2}{*}{\textbf{Methods}} & \multicolumn{2}{c}{\textbf{Llama3.1-8B}} & \multicolumn{2}{c}{\textbf{Mistral-7B-v0.3}} & \multicolumn{2}{c}{\textbf{Qwen2.5-7B}} \\
\cline{2-7}
                               & \textbf{Base} & \textbf{Instruct} & \textbf{Base} & \textbf{Instruct} & \textbf{Base} & \textbf{Instruct} \\
\hline
\textbf{AvgSTL}            & 65.43 & 65.49 & 65.77 & 65.74 & 64.34 & 64.67 \\
\textbf{MTL}                 & \textbf{65.84} & 65.54 & 65.43 & 65.06 & 64.62 & 64.77 \\
\textbf{SEE(1\%)}             & 65.36 & \textbf{65.59} & \textbf{65.80} & 65.85 & 64.68 & \textbf{64.94} \\
\textbf{SEE(10\%)}            & 65.68 & 65.55 & 65.76 & \textbf{65.86} & \textbf{64.85} & 64.80 \\
\hline
\end{tabular}
\caption{AR results of the latest models on ten SuperNI tasks.}\label{tab:latest_10}
\end{table*}

\section{Impact of \(\tau\) on SEE}

\begin{figure}[!h]
    \centering
    \includegraphics[width=0.8\linewidth]{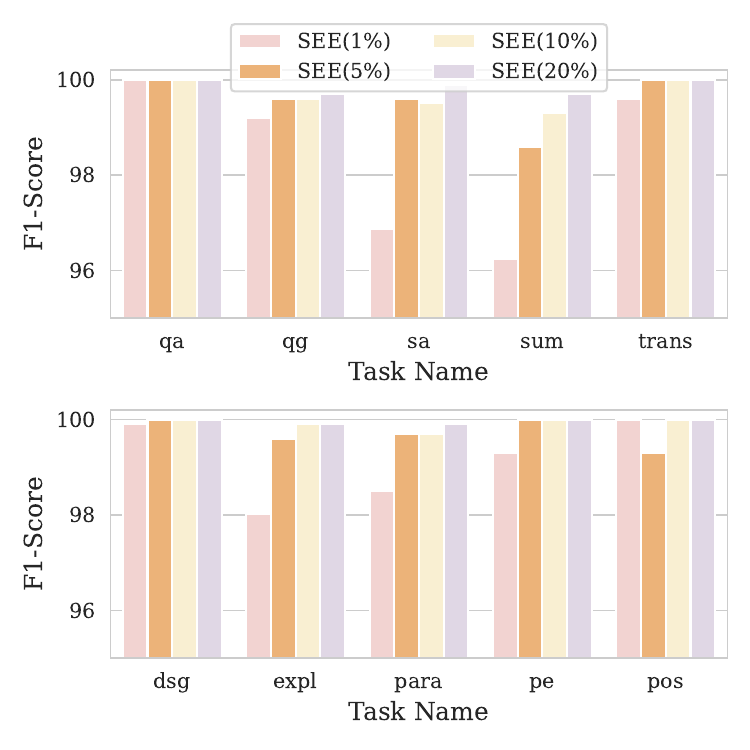}
    \caption{Details of the F1-score for each task on 10 SuperNI tasks.}
    \label{fig:route-analysis}
\end{figure}

We conduct a comprehensive evaluation of the influence of various values of \(\tau\) on the performance of our SEE framework across 10 distinct SuperNI tasks. The detailed routing F1-scores for each model on every task are presented in Figure~\ref{fig:route-analysis}, offering a thorough analysis of the model's performance under different configurations.

\label{app:routing_analysis}

\section{Extra Overhead Analysis}
\label{app:extra_overhead}
Although the increase in the number of tasks introduces additional computational overhead, this extra cost is still affordable. To demonstrate this, we calculate the additional latency of the SEE framework compared to LLMs. The latency of LLMs is given by:

\[
Latency_0 = TTFT + TPOT \times N
\]

where:
\begin{itemize}
    \item \textbf{TTFT} is the \textbf{Time to First Token} (the initial prefix latency),
    \item \textbf{TPOT} is the \textbf{Time Per Output Token} (the average time per output token),
    \item \textbf{N} is the number of output tokens.
\end{itemize}

In the SEE framework, extra latency is introduced due to multiple prefixing operations. Suppose we have \textbf{M} experts, labeled as 1, 2, \dots, M. Assuming the probability of a query hitting any expert or being out-of-distribution (OOD) is uniform, we define \textbf{X} as the random variable representing the query routing results.

\[
\begin{array}{cccccc}
\hline
X & 1 & 2 & \dots & M & \text{Exception} \\
\hline
P & \frac{1}{M+1} & \frac{1}{M+1} & \dots & \frac{1}{M+1} & \frac{1}{M+1}  \\
\hline
\end{array}
\]

To route to $Expert_i$, $i$ prefixing operations are required. Additionally, exceptions may occur in each expert, which causes the expected number of routing steps to be $(1 + M) / 2$. Therefore, the expected number of routing steps for the SEE framework can be calculated as:

\[
E[X] = \sum_{i=1}^{M} \frac{1}{M+1} \times i + \frac{1}{M+1} \times \frac{1 + M}{2}
\]

Simplifying this, we get:

\[
E[X] = \frac{M}{2} + \frac{1}{2} = \frac{M + 1}{2}
\]

Thus, the latency for the SEE framework is:

\[
Latency_1 = TTFT \times \frac{M+1}{2} + TPOT \times N
\]

Now, we can calculate the extra overhead as:

\[
ExtraOverhead = \frac{Latency_1 - Latency_0}{Latency_0}
\]

Substituting the expressions for \(Latency_1\) and \(Latency_0\), we get:

\[
ExtraOverhead = \frac{TTFT \times \frac{M-1}{2}}{TTFT + TPOT \times N}
\]

This simplifies to:

\[
ExtraOverhead = \frac{1}{2} \times \frac{1}{1 + \frac{N}{M-1} \times \frac{TPOT}{TTFT}}
\]

As shown in~\cite{zhong2024distservedisaggregatingprefilldecoding}, when serving an LLM with 13B parameters under a synthetic workload with input length = 512 and output length \textbf{N = 64} on an NVIDIA 80GB A100, the ratio \textbf{TPOT/TTFT} approximates \textbf{1/10}. Assuming we have \textbf{M = 10} experts, the extra overhead is:

\[
ExtraOverhead = \frac{1}{2} \times \frac{1}{1 + \frac{64}{10-1} \times \frac{1}{10}} = 0.29
\]

Even with a long input prefix and a short response, this results in only $0.29$ additional latency compared to traditional LLMs.

When the output length increases, this latency can be reduced to a very low level. For example, when the output length is equal to the input length (e.g., 512 tokens), the ratio of $TPOT/TTFT$ exceeds $1/10$, and the additional overhead is:
\[
ExtraOverhead < \frac{1}{2} \times \frac{1}{1 + \frac{512}{10-1} \times \frac{1}{10}} = 0.075
\]

\begin{table*}[h]
\centering
\small
\begin{tabular}{lccc}
\toprule
\textbf{Abbr.} &
\textbf{Category} &
\textbf{Name} &
\textbf{NLU task} \\
\midrule
QA & Question Answering & task024\_cosmosqa\_answer\_generation & - \\
QG & Question Generation & task074\_squad1.1\_question\_generation & - \\
SA & Sentiment Analysis & task1312\_amazonreview\_polarity\_classification & + \\
Sum. & Summarization & task511\_reddit\_tifu\_long\_text\_summarization & - \\
Trans. & Translation & task1219\_ted\_translation\_en\_es & - \\
DSG & Dialogue Sentence Generation & task574\_air\_dialogue\_sentence\_generation & - \\
Expl. & Explanation & task192\_hotpotqa\_sentence\_generation & - \\
Para. & Paraphrasing & task177\_para-nmt\_paraphrasing & - \\
POS & POS Tagging & task346\_hybridqa\_classification & + \\
PE & Program Execution & task064\_all\_elements\_except\_first\_i & - \\
\bottomrule
\end{tabular}
\caption{Details of the selected 10 SuperNI tasks.}
\label{tab:task-info} 
\end{table*}
\end{document}